\documentclass[10pt,twocolumn,letterpaper]{article}

\usepackage{cvpr}
\usepackage{times}
\usepackage{epsfig}
\usepackage{graphicx}
\usepackage{amsmath}
\usepackage{amssymb}
\usepackage{romannum}
\usepackage{verbatim}
\usepackage{multirow}
\usepackage{adjustbox}
\usepackage{subcaption}
\usepackage{caption}

\usepackage[breaklinks=true,bookmarks=false]{hyperref}

\cvprfinalcopy 

\def\cvprPaperID{****} 

\begin{document}

\title{Multi-Expert Gender Classification on Age Group by Integrating Deep Neural Networks}

\author{Jun Beom Kho\\
Yonsei University\\
50 Yonsei-ro, Seodaemun-gu, Seoul 03722, Republic of Korea.\\
{\tt\small kojb87@hanmail.net}
}

\maketitle

\begin{abstract}
    Generally, facial age variations affect gender classification accuracy significantly, because facial shape and skin texture change as they grow old.~This requires re-examination on the gender classification system to consider facial age information.~In this paper, we propose Multi-expert Gender Classification on Age Group (MGA), an end-to-end multi-task learning schemes of age estimation and gender classification.~First, two types of deep neural networks are utilized; Convolutional Appearance Network (CAN) for facial appearance feature and Deep Geometry Network (DGN) for facial geometric feature.~Then, CAN and DGN are integrated by the proposed model integration strategy and fine-tuned in order to improve age and gender classification accuracy.~The facial images are categorized into one of three age groups (young, adult and elder group) based on their estimated age, and the system makes a gender decision according to average fusion strategy of three gender classification experts, which are trained to fit gender characteristics of each age group.~Rigorous experimental results conducted on the challenging databases suggest that the proposed MGA outperforms several prior arts with smaller computational cost.
\end{abstract}

\section{Introduction}
\label{introduction}
For realistic automated gender classification system, facial images taken in unconstrained environmental conditions (e.g., CCTV and video cameras) have to be considered. This is an important challenge for the gender classification in the wild for obvious reasons such as low image quality caused by blurring, illumination variations and low resolution that can severely affect the system performance. In recent years, due to the success of Deep Neural Networks (DNN) in pattern recognition area, the gender classification has applied DNN and showed the promising results in the various conditions \cite{Antipov2016, Levi2015, Zhu2016, Han2018, Zhang2017, RODRIGUEZ2017563}.
\begin{figure}[!h]
\begin{center}
    \includegraphics[width=1.0\linewidth]{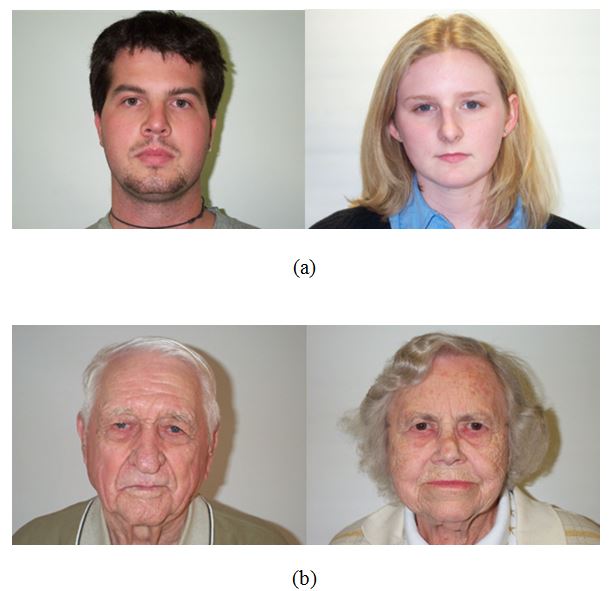}
\end{center}
\caption{Facial images of males and females of different ages in the PAL database \cite{Minear2004}}
\label{fig:figure_1}
\end{figure}
However, besides above challenging environmental conditions, the performance of the gender classification is affected by facial aging effect, which changes the discriminative facial features over age.~In general, the facial features characterizing a gender change as a person advances in age.~In Fig.~\ref{fig:figure_1}~(a), it is clear that the skin difference between the male and female can be prominently attributed to factors such as skin color, existence of beard, and thickness of eyebrows.~However, as the number and size of wrinkles and dark spots on faces increase as a person ages, the distinction in facial appearance between male and female becomes less obvious (see Fig.~\ref{fig:figure_1}~(b)).~The empirical study of facial aging effect to the gender classification was shown in Guo \etal \cite{Guo2009}, which reported that the gender classification performance deteriorates for young and senior age groups. Although the previous gender classifications using Convolutional Neural Network (CNN) have shown prominent classification accuracy in low quality images captured under unconstrained conditions, the negative effect of facial aging to the gender classification has not been considered yet.~Therefore, to overcome this issue, we propose deep gender classification method composed of multi-experts specialized in different age groups.~Each expert is constructed by combining two deep networks; the first one, referred as Convolutional Appearance Network (CAN), is CNN implemented to focus on facial appearance feature, and the second network, Deep Geometry Network (DGN), is a DNN designed for facial shape feature.~In the CAN, the lightweight CNN is utilized, because the gender classification does not require complex and deep architecture.~The second network DGN is proposed to extract the discriminative facial shape features robust to facial pose variation due to utilization of half of face. Through the model integration strategy, CAN and DGN are joint fine-tuned to improve overall performance.

In the previous gender classification researches (not deep learning based), the facial geometric features have been utilized for both age and gender classification \cite{Ferrario1993, Makinen2008, Samal2007, Xu2008}, and the performance was improved when they are incorporated with the facial appearance features.~Furthermore, apart from facial appearance features, the facial shape changes subtly after young age; thus facial shape feature can be exploited as a signature to resolve facial aging effect of the gender classification.~To the best of our knowledge, the facial geometric feature has not been adopted in the previous DNN-based gender classification researches, so the further performance improvement is expected when the proposed DGN is utilized.

The proposed method, referred as multi-expert gender classification on age group (MGA), is not limited to only gender classification, but also designed to predict age of facial image simultaneously.~The main purpose of MGA is focused on improving the performance of the gender classification, but the proposed method is exploited for joint age estimation and gender classification.~In general, multi-task learning schemes for facial age and gender classifications are based on CNN architecture which shares some layers and achieve better performance than training a single task one by one \cite{Han2018, Zhu2016}.~However, they are nothing more than sharing same layers without any compensation among correlated tasks.~Different from the previous joint age estimation and gender classification, the proposed MGA utilizes the result of age prediction to make gender decision.

Therefore, the major contributions of this paper are as follows:
\begin{enumerate}
    \item The relation between facial age and gender information is leveraged by the proposed MGA composed of gender classification experts on age groups.
    \item Deep learning-based facial geometric feature for age estimation and gender classification is investigated via the proposed DGN, and the test results prove that the DGN is beneficial to performance improvement.
    \item The model integration strategy of CAN and DGN is proposed to combine facial appearance and geometric features without enlarging model size. 
    \item The extensive experiments on datasets (Adience, Morph-\Romannum{2}) show that the proposed MGA outperforms the state-of-the-art methods with smaller computational cost, and discriminative facial area for the gender classification is analyzed.
\end{enumerate}

The remainder of this paper is organized as follows:~Section 2 illustrates the related works.~Section 3 describes the architecture of CAN, DGN and the proposed integrated network.~Section 4 explains the architecture of the proposed multi-expert gender classifications on age groups.~Section 5 presents our experimental results.~Finally, Section 6 concludes by summarizing and outlining future works.

\section{Related Works}
\label{Related Works}

Several studies have shown that humans can determine a person'’s gender easily and accurately using only facial information \cite{Brown1993, Burton1993}.~However, this is not a trivial task for machine-based gender recognition systems and remains as a challenge to date.~In the search for effective and robust solutions to this problem, most reported methods have been developed around the notion of geometry-based and appearance-based approaches \cite{Shih2013}.~Geometry-based approaches use the distance or ratio among the facial feature points, whereas appearance-based approaches extract the facial skin texture and edge information.

The geometry-based approaches primarily utilize eyebrow thickness, mouth width, thickness of lips, or facial shape as representative features.~These features are typically obtained by calculating the distances between facial feature points, and dubbed geometric features.~For instance, Xu \etal \cite{Xu2008} used Active Appearance Model to locate 83 facial feature points, and selected the ten most beneficial distances for gender classification.~Ferrario \etal \cite{Ferrario1993} and Samal \etal \cite{Samal2007} further studied various distances between selected facial feature points in order to determine the most useful geometric features for classifying gender.~Makinen and Raisamo \cite{Makinen2008} showed how the alignment of the face, which includes face normalization and image resizing, affects gender classification performance by adopting the face alignment to existing gender classification methods.~The geometry-based approach has merits such as light computation, robustness to blurriness, noise and illumination variations.~However, its performance relies on the accuracy of facial feature point detection.

The appearance-based methods use skin texture and the edge information of a facial image, and this approach is widely utilized in the recent gender classifications.~Ramesha \etal \cite{Ramesha2012} studied the distribution of pixel intensity values across certain facial regions in order to differentiate between male and female faces.~The result revealed that males and females have different texture patterns in their faces.~Shan \cite{Shan2012} combined the Local Binary Patterns (LBP) with AdaBoost, known as BoostLBP, to eliminate redundant information related to the texture features.~Guo \etal \cite{Guo2013, Guo2014} extracted Biologically-inspired features (BIF) from facial images, and minimized the feature dimension using Canonical Correlation Analysis (CCA) or Partial Least Squares (PLS).~The systematic comparison between CCA and PLS showed that the CCA outperforms the PLS with respect to running time and error rate.~An automated demographic (age, gender, and race) estimator is presented in Han \etal \cite{Han2015} that utilizes BIF for feature representation and hierarchical classifier for prediction.~Modesto \etal \cite{Modesto2016} showed that periocular region provides reliable information for the gender classification, and it is useful when the whole face is not visible.~Jia \etal \cite{Jia2015} proposed online classifiers which are learned using an ensemble of linear classifiers with four million images, and proved the positive effect of large training sets in gender classification tasks.~Azarmehr \etal \cite{Azarmehr2015} proposed a segmental dimensionality reduction technique to reduce computational requirements and memory, and the results showed that the methodology was a viable choice for real-time gender classification.~In general, even though these appearance-based methods perform satisfactory, they are not robust to blurriness, noise and illumination variation.

Recently, CNNs for unconstrained gender classification have been studied, and they reported that the performance of the CNN-based gender classification method outperformed the previous methods.~Antipov \etal \cite{Antipov2016} proposed minimal CNN architecture which needs less training data and time, and showed that ensemble of 3 CNNs improved the gender classification accuracy.~Levi and Hassner \cite{Levi2015} showed that the robust age and gender classifications under unconstrained conditions can be built by using a simple CNN architecture.~Zhu \etal \cite{Zhu2016} proposed joint age and gender recognition using simple CNN architecture, and showed that multi-task learning achieves better performance than learning a single task.~Mansanet \etal \cite{Mansanet2016} proposed a new model called Local Deep Neural Network for the gender classification.~Facial local patches are extracted, and a simple voting is performed to take into account the contribution of all patches driven by DNN.~Qawaqneg \etal \cite{Qawaqneh2017} proposed the age and gender classification, which consists of two different DNNs trained on different feature sets.~They showed that jointly fine-tuned DNNs improve the accuracy in the age and gender classification.~Xing \etal \cite{Xing2017} proposed hybrid multi-task learning architecture to consider race and gender information for improving the age estimation accuracy.~The experimental results showed that the hybrid multi-task learning outperformed the state-of-the art methods.~Han \etal \cite{Han2018} proposed heterogeneous facial attribute estimation using multi-task learning.~Multiple facial attributes were estimated from a single face by tackling attribute correlation and heterogeneity with CNN.~For CNN-based gender classification, large datasets acquired under uncontrolled conditions such as The labeled Faces in the Wild (LFW), Adience and MORPH have been used to evaluate the performances.

\section{Proposed Integrated Network}
As mentioned in Section~\ref{Related Works}, both facial appearance and geometric features were used for the facial gender classification.~However, recent DNN-based gender classification researches utilize only appearance-based features, because they focused on improving CNN architecture.~In this paper, the simple DGN, which has small amount of parameters, is proposed, and the proposed model integration strategy of CAN and DGN improves the performance of both age and gender classification.~In this section, CAN and DGN are introduced, and the model integration between CNN and DGN are explained. 

\subsection{Preprocessing}
\begin{figure}[!h]
\centering
\begin{subfigure}[t]{0.4\linewidth}
    \centering
    \includegraphics[width=1.1\linewidth]{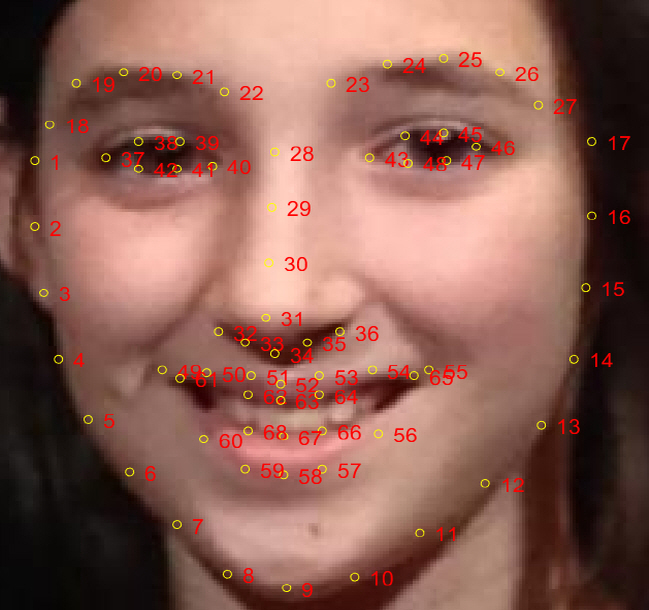}
    \caption{}
    \label{fig:figure_2a}
\end{subfigure}
\quad
\begin{subfigure}[t]{0.4\linewidth}
    \centering
    \includegraphics[width=1.1\linewidth]{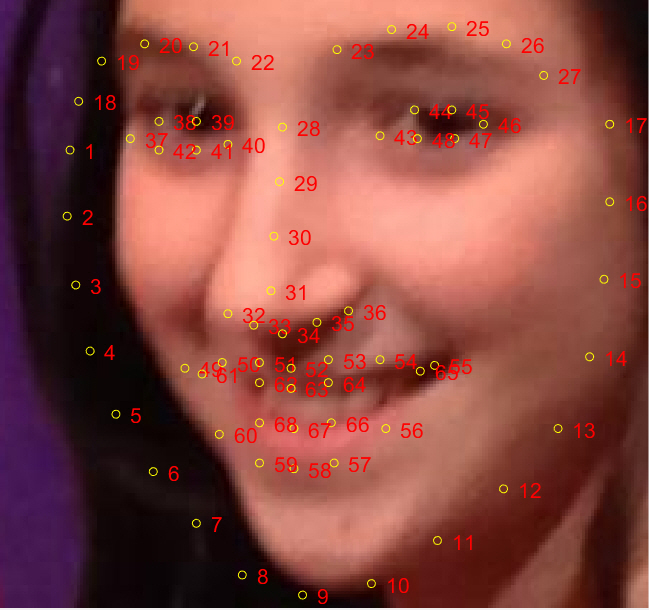}
    \caption{}
    \label{fig:figure_2b}
\end{subfigure}
\caption{68 facial landmarks of facial images from Adience database \cite{Eidinger2014}}
\label{fig:figure_2}
\end{figure}

The face detection, facial pose estimation and facial landmark detection are performed by using open source Dlib library \cite{dlib09} in this paper.~From detected face images, the Dlib library located 68 facial landmarks as shown in Fig.~\ref{fig:figure_2}, and the obtained facial landmarks are used for image alignment and the geometric feature extraction.~The rotation of the face is normalized by making the inter-ocular degree parallel.

\subsection{Convolutional Appearance Network}
\begin{figure}[!h]
\begin{center}
    \includegraphics[width=1.0\linewidth]{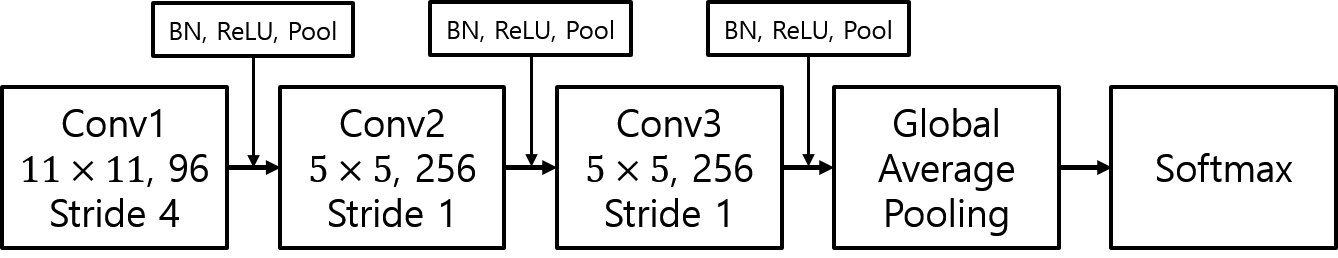}
\end{center}
\caption{CAN architecture}
\label{fig:figure_3}
\end{figure}

The lightweight CNN in \cite{Xing2017} is selected and modified for the baseline architecture of CAN.~As shown in Fig.~\ref{fig:figure_3}, three layers with 96, 256 and 384 filters are sequentially used for CAN. However, different from \cite{Xing2017}, Batch Normalization (BN) \cite{Ioffe2015} is adopted between convolution layer and activation function.~The Rectified Linear Unit (ReLU) \cite{Nair2010} is used for the activation function of all layers, followed by a $3\times3$ max pooling layer with a stride 2.~Furthermore, instead of two fully-connected layers used in \cite{Xing2017}, Global Average Pooling (GAP) is used to minimize overfitting by reducing the total number of parameters \cite{DBLP:journals/corr/LinCY13}.~Although this architecture is simple, it can show prominent performance with short processing time and small number of parameters.

\label{CAN}

Two evaluation metrics of CAN are used for the age estimationand gender classification.~Mean Absolute Error (MAE) is the most common evaluation metric for the age estimation, and the previous research \cite{Xing2017} reported that MAE showed the best performance compared to other metrics for the age estimation.~Thus, MAE is used for the age estimation of CAN, and it can be defined as
\begin{equation}
\label{eq1}
\mathcal{L}_{a}^{CAN}=\frac{1}{N}\sum_{i=1}^N\lvert y_{i}^a-\Tilde{y}_{i}^a\rvert,
\end{equation}
where $N$ is the total number of sample images, $y_i$ and $\Tilde{y}_i$ are ground truth age and predicted age of $i$th image respectively. By using MAE, the age estimation of CAN is formulated as regression problem.

The gender classification is binary selection problem. Thus, cross entropy error function with softmax output is adopted for loss function as follows:
\begin{equation}
\label{eq2}
\mathcal{L}_{g}^{CAN}=-\frac{1}{N}\sum_{i=1}^N\sum_{c=1}^2y_{ic}^g\log(\Tilde{p}_{ic}^g),
\end{equation}
where  $N$ is the total number of sample images, $y^g_{ic}$ is ground truth label, $\Tilde{p}_{ic}^g$ is the softmax output of image $i$ with gender label $c$.~Therefore, we train CAN with the loss function $\mathcal{L}^{CAN}$ defined as follows:
\begin{equation}
\label{loss_CAN}
    \mathcal{L}^{CAN}=\mathcal{L}_a^{CAN}+\mathcal{L}_g^{CAN}.
\end{equation}

\subsection{Deep Geometry Network}
\label{section_DGN}
Generally, the facial geometric feature is robust under image translation and rotation when the facial landmarks are detected accurately.~However, if the facial pose is not frontal, facial landmarks will be twisted although locations of landmarks are accurate.~For example, as shown in Fig.~\ref{fig:figure_2b}, if a face is rotated to the right, right side of face is obscured and twisted. On the other hand, left side is less affected by facial pose.~Thus, we can still utilize the facial landmarks of less affected half of face (e.g.\ right side is used when face is rotated to the left and vice versa).~In DGN, each face is divided into two halves, and only one side, which is less twisted by facial pose is used.~However, we cannot use the facial landmark positions directly for the input of DGN, because left and right facial landmarks have opposite positions, which makes input vector different according to facial pose.~Thus, in this paper, the facial landmarks in left side of face are projected into right facial positions on the basis of nose.

After that, scale normalization is performed for the extracted facial landmarks to eliminate the scale variations caused by distance between face and camera.~Generally, the facial scale normalization is carried out by equalizing the distance among two eyes \cite{Chen2006}, but, in DGN, distance between nose and eye is used instead, because only a half of face is utilized ($28_{th}$ and $40_{th}$ landmark points are used for the right face, and $28_{th}$ and $43_{th}$ are used for the left face in Fig.~\ref{fig:figure_2}). 

Let $\mathbf{x}=\{x_1,y_1,x_2,y_2,\ldots,x_n,y_n \}$ be a vector, which contains coordinates of extracted facial landmarks from one image, where $n$ is the number of points in a half of face, $x_j$ and $y_j$ are xy-coordinates of $j_{th}$ facial landmark.~When nose position ($31_{th}$ facial landmark in Fig.~\ref{fig:figure_2}) is set as a basis point, x,y coordinates of facial landmarks are normalized as follows:
\begin{equation}
\label{eq4}
    \Bar{x}_i=\frac{x_{i}-x_{nose}}{\sigma_x},\qquad\Bar{y}_i=\frac{y_{i}-y_{nose}}{\sigma_y},
\end{equation}
where $x_{nose}$ and $y_{nose}$ are xy-coordinates of nose point, $\sigma_x$ and $\sigma_y$ are standard deviations of all facial landmark positions from a half of face.~Thus, a vector $\Bar{\mathbf{x}}=\{\Bar{x_1},\Bar{y_1},\Bar{x_2},\Bar{y_2},\ldots,\Bar{x_n},\Bar{y_n} \}$ is obtained from Eq.~(\ref{eq4}).~In addition, DGN also utilizes Euclidean distances among facial feature points to classify age and gender of face, the distances are obtained as follows:
\begin{equation}
\label{eq5}
    d_{i,j}=\sqrt{\left(\Bar{x}_i-\Bar{x}_j\right)^2+\left(\Bar{y}_i-\Bar{y}_j\right)^2},
\end{equation}
where $d_{i,j}$ is the Euclidean distance between $i_{th}$ and $j_{th}$ facial landmarks, and the obtained distances are added to the vector $\Bar{\mathbf{x}}$.~Eventually, DGN takes an input vector of $\Hat{\mathbf{x}}=\{\Bar{x}_1,\Bar{y}_1,\Bar{x}_2,\Bar{y}_2,\ldots,\Bar{x}_n,\Bar{y}_n,d_{1,2},d_{1,3},\ldots,d_{n-1,n}\}$.
\begin{figure}[!h]
    \centering
    \includegraphics[width=0.8\linewidth]{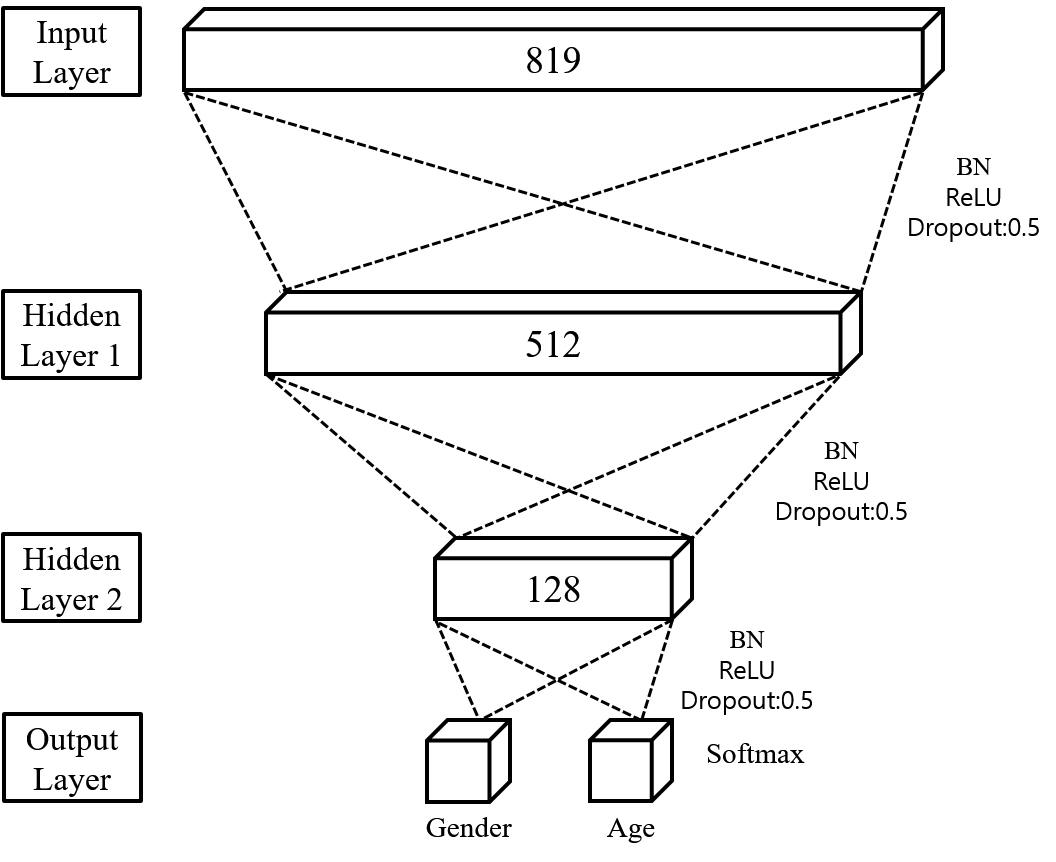}
    \caption{Deep Geometry Network}
    \label{fig:figure_4}
\end{figure}

As shown in Fig.~\ref{fig:figure_4}, two hidden layers are utilized in DGN, and the top layer is the softmax layer.~BN is also adopted to each hidden layer, and ReLU is used for the activation function of each hidden layer.~Same as CAN, a single DGN is jointly trained with gender and age loss functions for the multi-task learning.~The loss function $\mathcal{L}_g^{DGN}$ for the gender classification in DGN is same as Eq.~(\ref{eq2}), but Eq.~(\ref{eq1}) is not appropriate for the age estimation of DGN, because single geometric feature is not sufficient to estimate detail age of facial image.~Thus, we set DGN to be trained for age group classification instead of the age estimation, so the loss function is formulated as follows:
\begin{equation}
\label{eq6}
    \mathcal{L}_{ag}^{DGN}(x)=-\frac{1}{N}\sum_{i=1}^N\sum_{j=1}^ny_{ij}^{ag}\log(\Tilde{p}^{ag}_{ij}),
\end{equation}
where $N$ is the number of input images, $n$ is the number of age groups, $\Tilde{p}^{ag}_{ij}$ is the output probability value of softmax layer for $j$th age group label of $i$th image, $y_{ij}^{ag}$ is ground truth age group (we divide age range into 8 age groups e.g. 0-10, 11-20, \ldots, 70-79).~Finally, we train DGN with the loss function $\mathcal{L}^{DGN}$ defined as follows:
\begin{equation}
\label{loss_DGN}
    \mathcal{L}^{DGN}=\mathcal{L}_{ag}^{DGN}+\mathcal{L}_g^{DGN}.
\end{equation}

\subsection{Model Integration}
\label{model integration}
In DGN, the output of each layer can be represented as follows:
\begin{align}
\label{eq7}
    f_k^D(x_i)=ReLU(BN(W_k^Df_{k-1}^D(x_i)+b_k^D)),\nonumber\\ f_0^D(x_i)=\Hat{\mathbf{x}}, k=\{1,2,3\},
\end{align}
where $x_i$ is the input vector of each layer, BN is Batch Normalization, $W_k^D$ and $b_k^D$ are weight and bias parameters respectively, ~$k=\{1,2,3\}$ indicates hidden layer 1, hidden layer 2 and output layer respectively.~According to the previous gender classification researches, only a few selected distances are beneficial to classify gender of face \cite{Xu2008}.~Thus, how to select discriminative distances from facial landmarks is important for geometry-based gender classification.~DGN is trained to extract useful facial geometric features with the loss functions of $\mathcal{L}_{ag}^{DGN}$ and $\mathcal{L}_g^{DGN}$, so the output of hidden layer 2 $(f_2^D)$ can be used for the discriminative geometric feature.

In CAN, the feature maps from last convolutional layer are trained to correspond with categories (gender or age) by the spatial average of the global average pooling layer.~Therefore, in this paper, the output of the global average pooling layer in CAN, denoted as $f^C(x'_i)$ ($x'_i$ is an input image), is used and combined with $(f_2^D)$.

\begin{figure}[!h]
    \centering
    \includegraphics[width=1.0\linewidth]{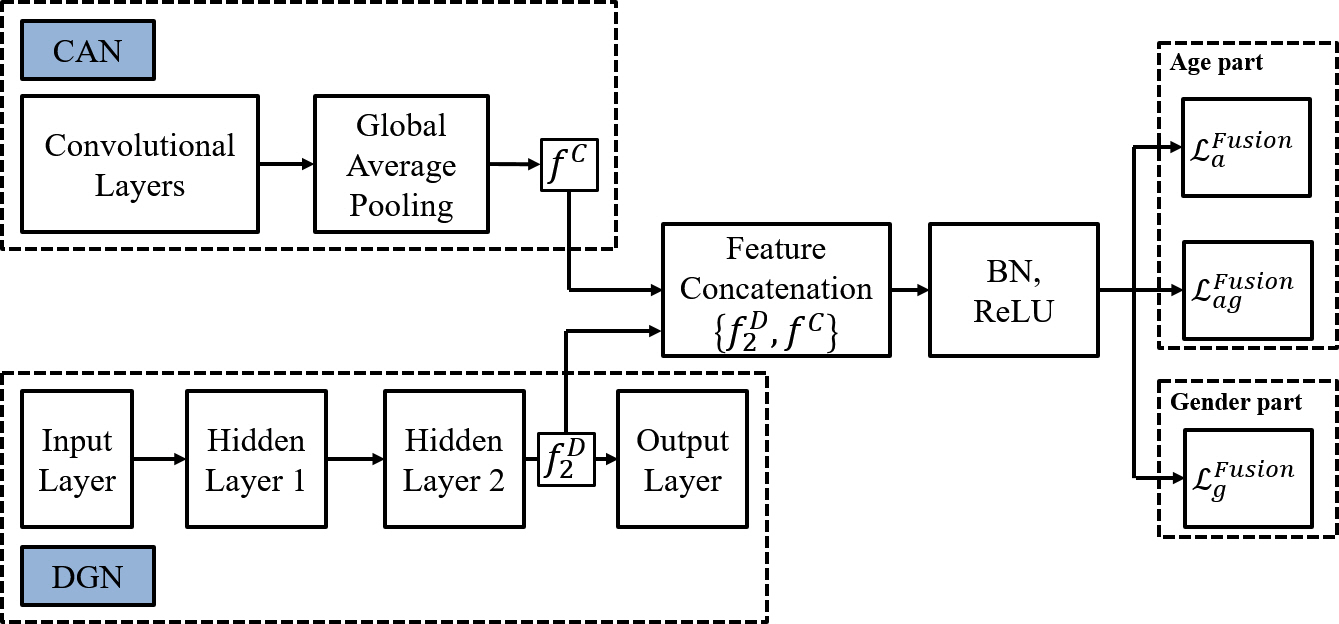}
    \caption{Integrated network of CAN and DGN}
    \label{fig:figure_5}
\end{figure}

Let $f_2^D(x_i)$ and $f^C(x'_i)$ be $N_D$ and $N_C$ dimensional vectors respectively;~then two vectors are concatenated into a vector $\boldsymbol{f}=\{f_2^D(x_i), f^C(x'_i)\}\in\mathbb{R}^{N_D+N_C}$ and connected to top layer for the final decision.~The proposed integrated network (IN) of CAN and DGN is dispicted in Fig.~\ref{fig:figure_5}. 

From pre-trained CAN and DGN, we retrain the IN with $\mathcal{L}^{fusion}$ defined as follows: 
\begin{align}
\label{loss_fusion}
    \mathcal{L}^{fusion}=\mathcal{L}_g^{fusion}+\alpha_1\mathcal{L}_a^{fusion}+\beta_1\mathcal{L}_{ag}^{fusion}\nonumber\\
    =-\frac{1}{N}\sum_{i=1}^N\sum_{c=1}^2y_{ic}^g\log(\Tilde{p}_{ic}^g)+\alpha_1\frac{1}{N}\sum_{i=1}^N\lvert y_i^{a}-\Tilde{y}_i^a\rvert\nonumber\\-\beta_1\frac{1}{N}\sum_{i=1}^N\sum_{j=1}^3y_{ij}^{ag}\log(\Tilde{p}^{ag}_{ij}),
\end{align}
where $\mathcal{L}^{fusion}_g$ is implemented for the gender classification, $\mathcal{L}^{fusion}_a$ and $\mathcal{L}^{fusion}_{ag}$ are used for the age and age group classification respectively, $\alpha_1$ and $\beta_1$ are tuning parameters.~In Eq.~(\ref{loss_fusion}), the loss function of age group classification $(\mathcal{L}^{fusion}_{ag})$ is designed for three classes (young, adult and elder group) so that the input image is categorized into one of three age groups.

\begin{figure*}[!ht]
    \centering
    \includegraphics[width=.8\linewidth]{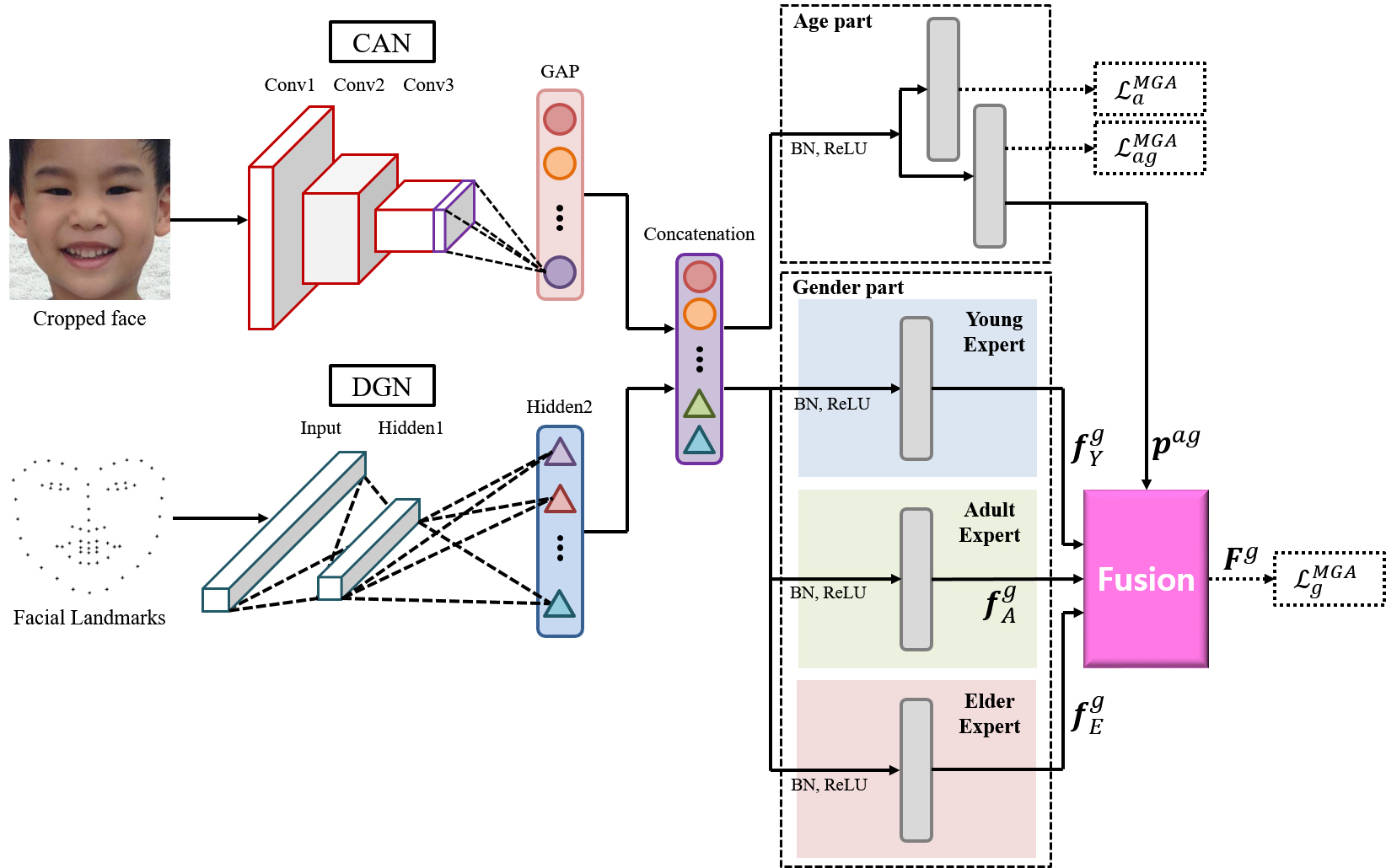}
    \caption{The architecture of the proposed MGA}
    \label{fig:figure_6}
\end{figure*}

\section{Multi-expert Gender Classification on Age Group}
\label{Multi-expert Gender Classification on Age Group}
As mentioned in Section~\ref{introduction}, the performance of the gender classification is affected by facial age, and this is mainly caused by facial feature change over age.~Thus, the discriminative features for gender classification are changed based on facial age, and the MGA is proposed to resolve this issue by constructing three gender classification experts trained in particular age groups. 

\subsection{Model Architecture}
As shown in Fig.~\ref{fig:figure_6}, the proposed MGA has two parts: age and gender parts.~The age estimation and age group classification are included in the age part, and the gender part is composed of three gender classification experts (young, adult and elder expert).~The architecture of the MGA is the extended version of IN that the gender part of IN is divided into three gender classifications by adding three expert layers. 

Let $\boldsymbol{p}^{ag}=[\Tilde{p}^{ag}_Y, \Tilde{p}^{ag}_A, \Tilde{p}^{ag}_E]$ be the output of the age group classification indicating probability of an input image belonging to each age group.~Then, the result of each gender classification expert is combined with the output of the age group classification by the average fusion strategy to get the final prediction ${F_c^{g}}$ defined as follows:
\begin{equation}
F_c^{g}=\sum_{k\in{\{Y,A,E\}}}\Tilde{p}^{ag}_kf^{g}_{k,c}
\end{equation}
where $c$ is the gender label and $f^{g}_{k,c}$ is the probability of an input image belonging to gender label $c$ obtained from $k$ expert.~Therefore, the final prediction to the gender label $c$ is made by $F_c^{g}$.

\subsection{Learning Network}
\label{Learning Network}
Four steps of training process are designed to train the proposed MGA:\\ 
1) CAN and DGN are pre-trained with $\mathcal{L}^{CAN}$ (Eq.~(\ref{loss_CAN})) and $\mathcal{L}^{DGN}$ (Eq.~(\ref{loss_DGN})) respectively.\\
2) IN is jointly fine-tuned from weights of CAN and DGN (step 1) with $\mathcal{L}^{fusion}$ when $\alpha_1=1$ and $\beta_1=1$ (Eq.~(\ref{loss_fusion})).\\
3) IN is fine-tuned three times with three age group images respectively to obtain weights of three expert layers (young, adult, elder expert) when other layers are frozen, $\alpha_1=0.0001$ and $\beta_1=0.0001$ (Eq.~(\ref{loss_fusion})).\\
4) the total MGA is retrained from end to end using weights obtained from step 1, 2 and 3 with $\mathcal{L}^{MGA}$ defined as follows:

\begin{align}
\label{loss function of MGA}
    \mathcal{L}^{MGA}=\mathcal{L}_g^{MGA}+\lambda_1\mathcal{L}_{ag}^{MGA}+\lambda_2\mathcal{L}_a^{MGA}\nonumber\\
    =-\frac{1}{N}\sum_{i=1}^N\sum_{c=1}^2y_{ic}^g\log(F_{ic}^g)+\lambda_1\frac{1}{N}\sum_{i=1}^N\lvert y_i^{a}    -\Tilde{y}_i^a\rvert\\-\lambda_2\frac{1}{N}\sum_{i=1}^N\sum_{j\in\{Y,A,E\}}y_{ij}^{ag}\log(\Tilde{p}^{ag}_{ij})\nonumber,
\end{align}
where $N$ is total number of images, $\mathcal{L}_{ag}^{MGA}=\mathcal{L}_{ag}^{fusion}$, $\mathcal{L}_{a}^{MGA}=\mathcal{L}_{a}^{fusion}$, $F_{ic}^{g}$ is the final gender prediction of $i_{th}$ image to the gender label $c$, $\lambda_1=0.1$ and $\lambda_2=0.1$.

\begin{figure}[!h]
    \centering
    \includegraphics[width=1.0\linewidth]{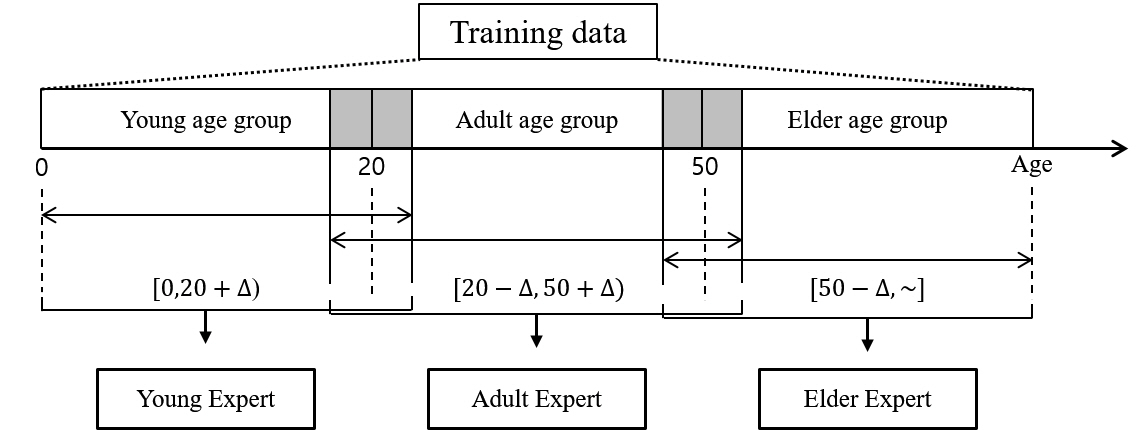}
    \caption{Training age ranges of gender classification experts}
    \label{fig:figure_7}
\end{figure}

In this paper, three age groups are seperated on the basis of 20 and 50 years (see Fig.~\ref{fig:figure_7}).~This is because prominent craniofacial shape change accurs under 20 years and facial skin ages significantly over 50 years \cite{Guo2009}.~From step 3, three experts are trained to fit facial gender characteristic of each age group, and their predictions are combined by the proposed MGA.

Therefore, high accuracy of the age estimation is essential to classify input images into the correct age group. However, considerable age estimation error is still reported in the state-of-the-arts (2$\sim$3 MAE), even though deep learning-based age estimation has been proposed.~Thus, misclassification of the input image to wrong age group (e.g.~young facial image is assigned to adult group) caused by the age estimation error will affect the performance of the proposed MGA.~This problem occurs more frequently for the images located close to age group boundary.~Therefore, to rectify this problem, the overlapping age region $(\Delta)$ is adopted to train each gender classification expert in the training stage (See Fig.~\ref{fig:figure_7}).~Thus, even though the age group of an input image is misclassified, the proposed MGA can compensate this problem.

\section{Experimental Results}
\subsection{Databases}
\label{Databases}
\begin{figure}[!h]
\centering
\begin{subfigure}[t]{\linewidth}
    \centering
    \includegraphics[width=0.8\linewidth]{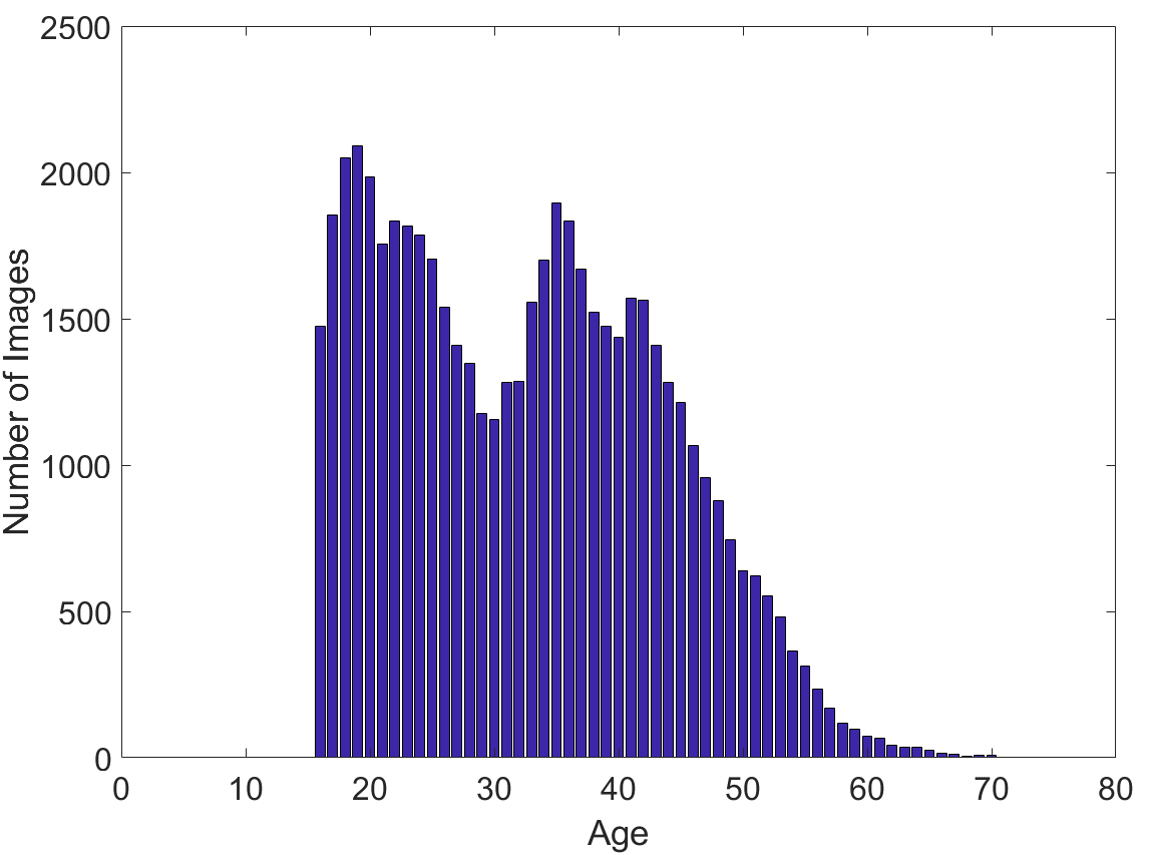}
    \caption{}
    \label{fig:figure_8a}
\end{subfigure}
\quad
\begin{subfigure}[t]{\linewidth}
    \centering
    \includegraphics[width=0.8\linewidth]{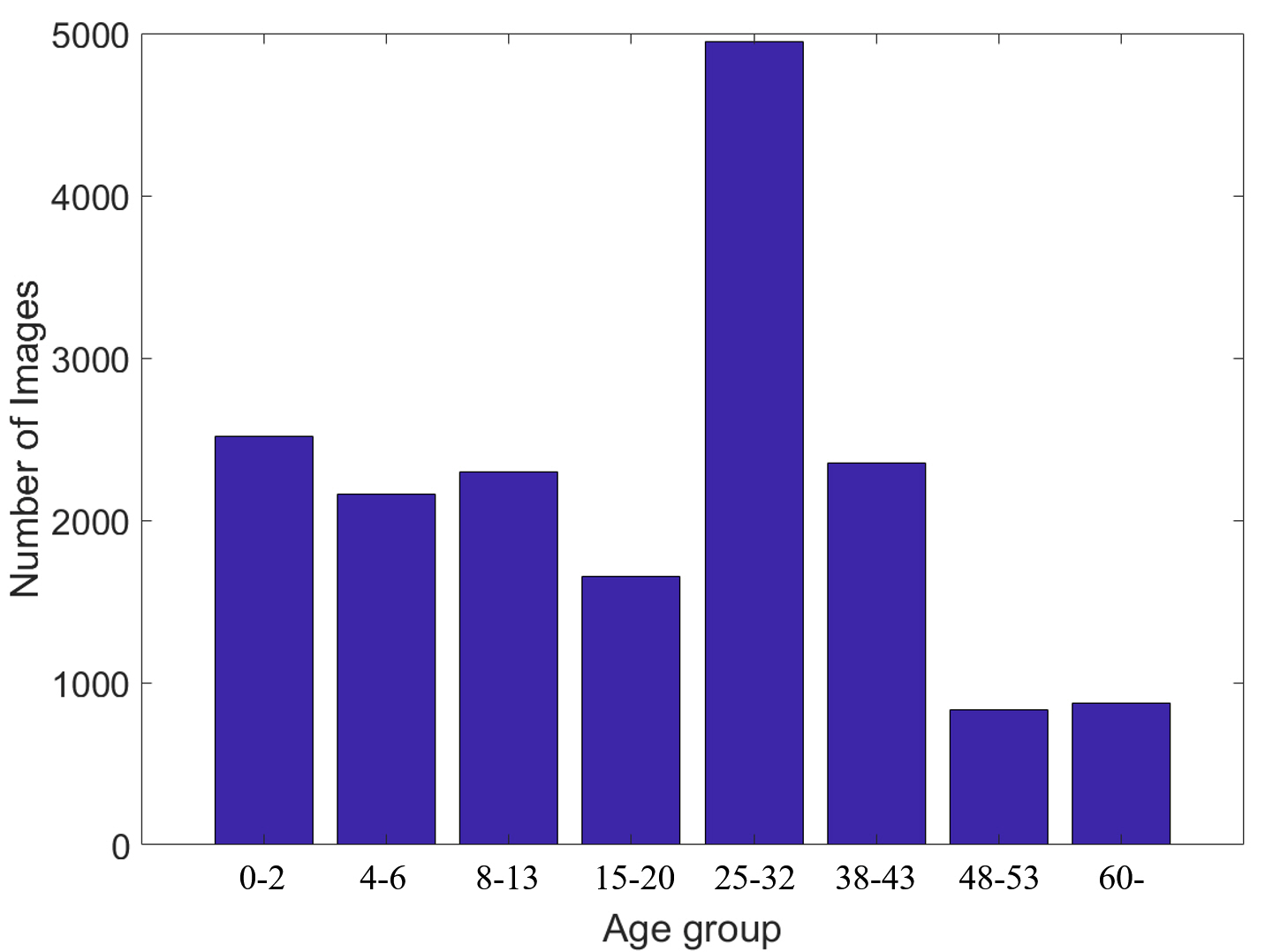}
    \caption{}
    \label{fig:figure_8b}
\end{subfigure}
\caption{Age distributions of datasets (a) MORPH-\Romannum{2} (b) Adience}
\label{fig:figure_8}
\end{figure}

Two datasets were used in this paper to show the performance of the proposed method: Adience, Morph-\Romannum{2}.~Adience~\cite{Eidinger2014}, one of the most challenging datasets, is used to show the performance of the proposed method under the unconstrained conditions.~Adience contains 26,580 images with both age group and gender information, and images were captured under variations of head pose, illumination condition and image quality.~There are 8 categories for age groups (i.e., 0-2, 4-6, 8-13, 15-20, 25-32, 38-43, 48-53, 60-), so only age group classification can be applied in Adience instead of the age estimation.~Morph-\Romannum{2} \cite{morph} contains about 55,000 face images with ages ranging from 16 to 77 years old.~The images obtained under relatively constrained conditions and less challenging pose variations are included.~Fig.~\ref{fig:figure_8} shows the age distributions of both datasets.

\subsection{Experimental settings}
The facial image is cropped and rescaled to $227\times227$ color image, and for data augmentation, each image is horizontally flipped and randomly rotated in $-40^{\circ}\sim40^{\circ}$ to be robust against the rotational change of images.~For the experiments, all models are trained and tested by Keras (Python code) \cite{chollet2015keras} with the TensorFlow backend on Geforce GTX 1080.~We use Adam optimizer, and each network is trained until the training loss is saturated.~Additional training details of networks are displayed in Table \ref{Network setting}.~All tests were conducted using 5-fold cross validation in which the provided folds are subject-exclusive.

\begin{table}[!h]
\centering
\caption{Network settings}
\label{Network setting}

\begin{adjustbox}{width=0.48\textwidth}
\begin{tabular}{lcccl}
\hline
Model & CAN & DGN & IN & \multicolumn{1}{c}{MGA} \\ \hline
Mini-batch size & 128 & 256 & 128 & 128 \\
Initial Learning rate & $10^{-2}$ & $10^{-2}$ & $10^{-3}$ & $10^{-3}$ \\
Learning rate decay & $5*10^{-4}$ & $5*10^{-4}$ & $5*10^{-4}$ & $5*10^{-4}$ \\
Activation function & Adam & Adam & Adam & Adam \\ \hline
\end{tabular}
\end{adjustbox}
\end{table}

\subsection{Performance Evaluation}

\subsubsection{Performance Analysis of the Integrated Network}
In this section, CAN, DGN and IN are evaluated to show the performance improvement when the proposed IN is utilized for the age estimation and gender classification.~On MORPH-\Romannum{2}, the age estimation accuracies of CAN and IN are evaluated by Mean Absolute Error (MAE), and DGN is tested in terms of age group classification with eight age groups (0$\sim$9, 10$\sim$19, $\ldots$, 70$\sim$79).~The performances of the age group classification are evaluated by exact (correct age group prediction) and 1-off (correct or adjacent age group prediction) \cite{Zhang2017}.~Table~\ref{CAN,DGN,IN on MORPH}~and~\ref{CAN,DGN,IN on Adience} show the results.

\begin{table}[!h]
\centering
\caption{The age estimation and gender classification accuracy of CAN, DGN and IN on MORPH-\Romannum{2}.}
\label{CAN,DGN,IN on MORPH}

\begin{adjustbox}{width=0.45\textwidth}
\begin{tabular}{p{3.8cm}lllllll}
\hline
\multirow{2}{*}{Method} & \multirow{2}{*}{Gender (\%)} & \multicolumn{2}{l}{Age} \\ \cline{3-4}
                        &        & MAE       & 1-off (\%)                        \\ \hline
CAN                     & 98.59      & 3.34            &    -                     \\
DGN                     & 88.28         & -            & 76.76                       \\
IN                      & \textbf{98.93}     & \textbf{3.05} & -                      \\ \hline
\end{tabular}
\end{adjustbox}
\end{table}

\begin{table}[!h]
\centering
\caption{The age estimation and gender classification accuracy of CAN, DGN and IN on Adience.}
\label{CAN,DGN,IN on Adience}

\begin{adjustbox}{width=0.45\textwidth}
\begin{tabular}{p{3.8cm}lllllll}
\hline
\multirow{2}{*}{Method} & \multirow{2}{*}{Gender (\%)} & \multicolumn{2}{l}{Age} \\ \cline{3-4}
                        &       & Exact (\%)     & 1-off (\%)                  \\ \hline
CAN                     &   90.31 &  70.63     &     86.21                     \\
DGN                     &    74.77 &  52.41         &   76.83                     \\
IN                      & \textbf{90.94}    &  \textbf{73.86}         & \textbf{87.57}                 \\ \hline
\end{tabular}
\end{adjustbox}
\end{table}

From Table~\ref{CAN,DGN,IN on MORPH}~and~\ref{CAN,DGN,IN on Adience}, all accuracies of DGN are worse than those of CAN.~However, the performances of age estimation and gender classification are improved when CAN and DGN are integrated by the proposed IN.~The result shows that DGN is beneficial to improve the overall performances of age estimation and gender classification, and this corresponds to the expectation that the performance improvement can be achieved when the geometric feature is utilized with facial appearance feature.~To the best of our knowledge, the facial geometric feature was not used for the previous deep learning-based gender classification and age estimation.~Thus, the investigation of deep learning-based geometry approach via DGN will contribute the gender classification and age estimation field.

\subsubsection{Performance Analysis on Age Groups}
As mentioned in Section \ref{introduction}, the facial features discriminating gender are changed as people age.~Thus, in this section, the effect of facial aging to the performance of the gender classification is analyzed, and the accuracy improvement due to the proposed gender classification expert is introduced. 

\begin{table}[!h]
\centering
\caption{The gender classification accuracy (\%) of three age groups on MORPH-\Romannum{2}.}
\label{gender classification of age groups on MORPH}

\begin{adjustbox}{width=0.45\textwidth}
\begin{tabular}{p{3.5cm}llrl}
\hline
\multirow{2}{*}{Method} & \multicolumn{3}{c}{Test age group} \\ \cline{2-4} 
                        & Young    & Adult    & Elder   \\ \hline
CAN                     & 98.18    & 98.69    & 98.05   \\
IN                      & 98.62    & 98.98    & 98.83   \\
IN (expert)             & \textbf{98.65}    & \textbf{99.06}    & \textbf{98.89}   \\ \hline
\end{tabular}
\end{adjustbox}

\end{table}
\begin{table}[!h]
\centering
\caption{The gender classification accuracy (\%) of three age groups on Adience.}
\label{gender classification of age groups on Adience}

\begin{adjustbox}{width=0.45\textwidth}
\begin{tabular}{p{3.5cm}llrl}
\hline
\multirow{2}{*}{Method} & \multicolumn{3}{c}{Test age group} \\ \cline{2-4} 
                        & Young    & Adult    & Elder   \\ \hline
CAN                     & 86.88    & 93.57    & 88.85   \\
IN                      & 87.80    & 94.01    & 89.93   \\
IN (expert)             & \textbf{88.13}    & \textbf{94.59}    & \textbf{91.77}   \\ \hline
\end{tabular}
\end{adjustbox}
\end{table}

Table~\ref{gender classification of age groups on MORPH}~and~\ref{gender classification of age groups on Adience} show the performance of gender classification on each age group.~IN (expert) indicates gender classification expert obtained by step 3 (explained in Section~\ref{Learning Network}), and CAN and IN are obtained by training all dataset at once.~From these experiments, we can notice that the gender classification accuracy is degraded when the test images are either young or elder group, and this proves that the facial aging affects the performance of the gender classification.~Thus, it is important to obtain better result on both young and elder age groups to improve the overall gender classification accuracy.

From Table~\ref{gender classification of age groups on MORPH}~and~\ref{gender classification of age groups on Adience}, IN (expert) achieves the best accuracy for all age groups on both MORPH-\Romannum{2} and Adience. The results support that the proposed gender classification expert is properly trained to utilize the characteristics of each age group.~The significant accuracy improvement by IN (expert) is achieved on Adience, especially elder age group, because the images in Adience are distributed across wide age ranges (see Fig.~\ref{fig:figure_8b}).~Thus, the proposed gender classification expert shows better efficacy on Adience dataset.~On the other hand, the classification accuracy is slightly improved on MORPH-\Romannum{2}, because age distribution of MORPH-\Romannum{2} is biased to the adult age group (see Fig.~\ref{fig:figure_8a}), which means the images are less affected by the facial aging.

\begin{figure}[!t]
\centering
\begin{subfigure}[t]{\linewidth}
    \centering
    \includegraphics[width=1\linewidth]{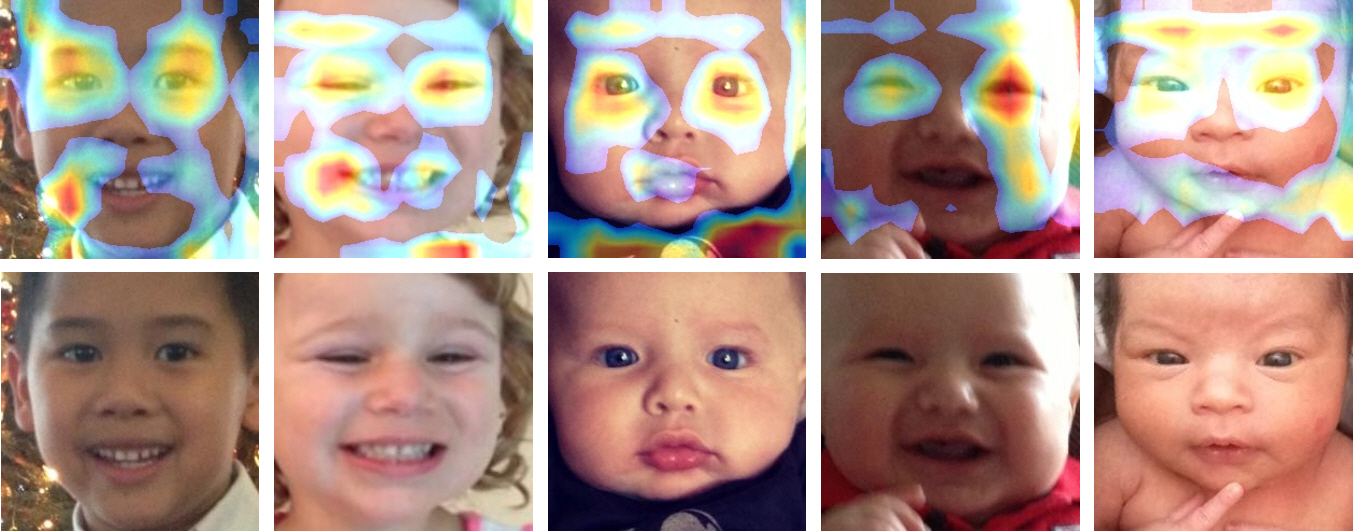}
    \caption{}
    \label{fig:figure_9a}
\end{subfigure}

\begin{subfigure}[t]{\linewidth}
    \centering
    \includegraphics[width=1\linewidth]{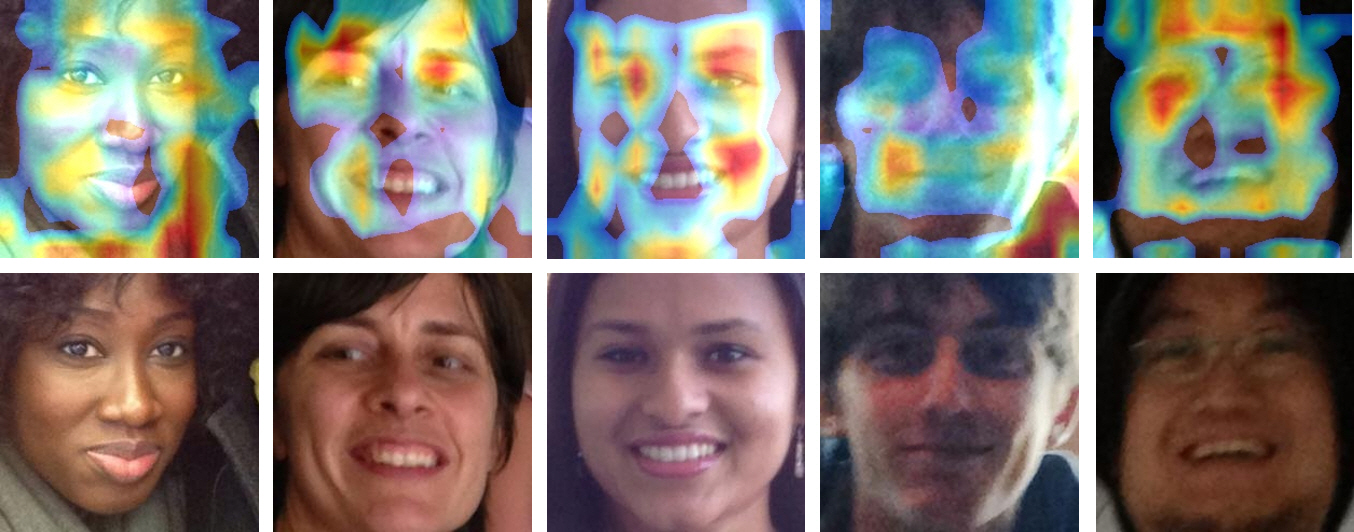}
    \caption{}
    \label{fig:figure_9b}
\end{subfigure}

\begin{subfigure}[t]{\linewidth}
    \centering
    \includegraphics[width=1\linewidth]{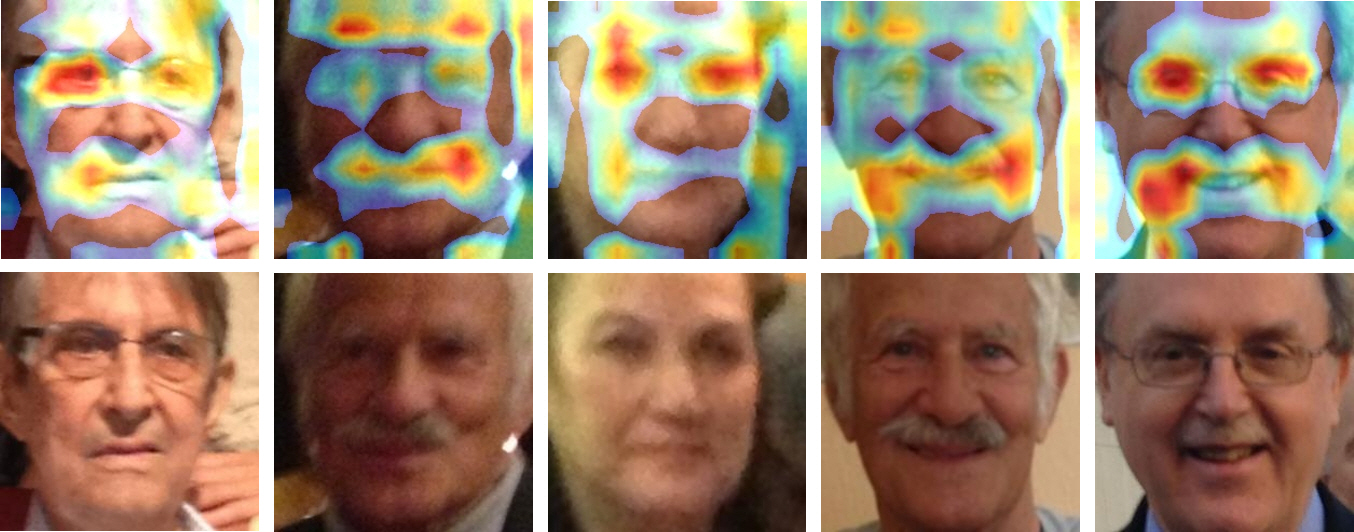}
    \caption{}
    \label{fig:figure_9c}
\end{subfigure}
\caption{CAMs of each expert gender classification from Adience (a) young, (b) adult, (c) elder.}
\label{fig:figure_9}
\end{figure}

\begin{figure}[!t]
\centering
\begin{subfigure}[t]{\linewidth}
    \centering
    \includegraphics[width=1\linewidth]{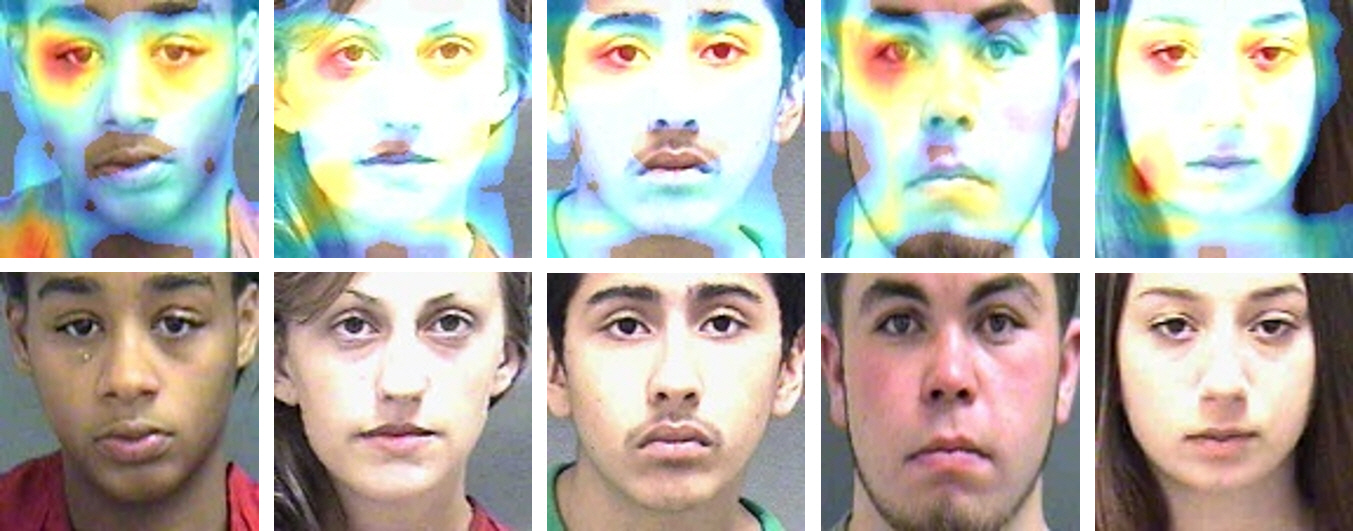}
    \caption{}
    \label{fig:figure_10a}
\end{subfigure}

\begin{subfigure}[t]{\linewidth}
    \centering
    \includegraphics[width=1\linewidth]{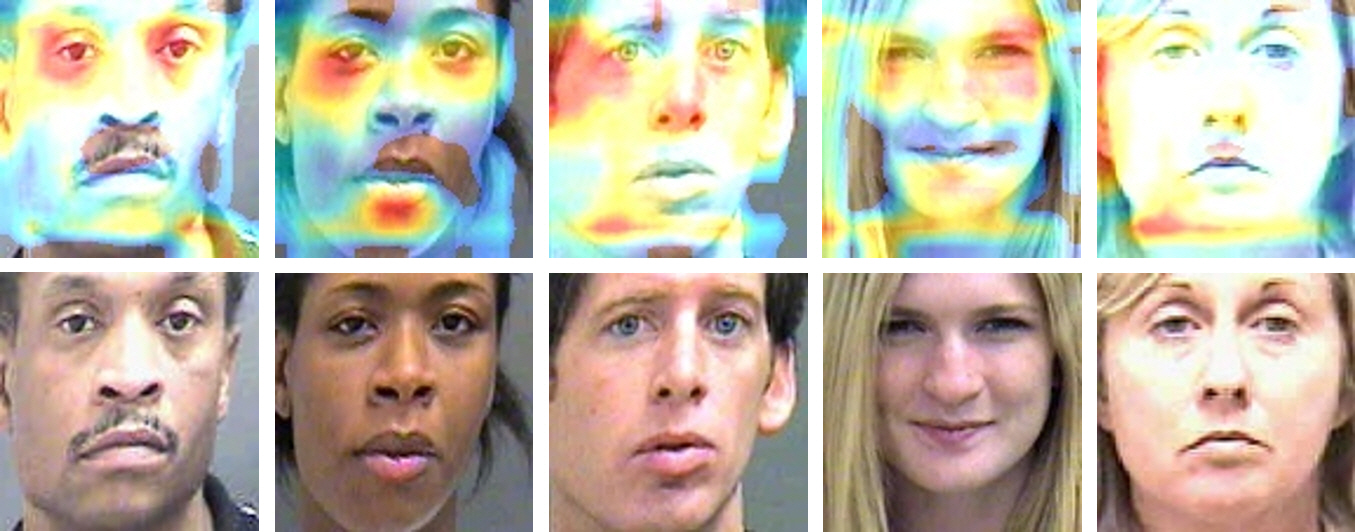}
    \caption{}
    \label{fig:figure_10b}
\end{subfigure}

\begin{subfigure}[t]{\linewidth}
    \centering
    \includegraphics[width=1\linewidth]{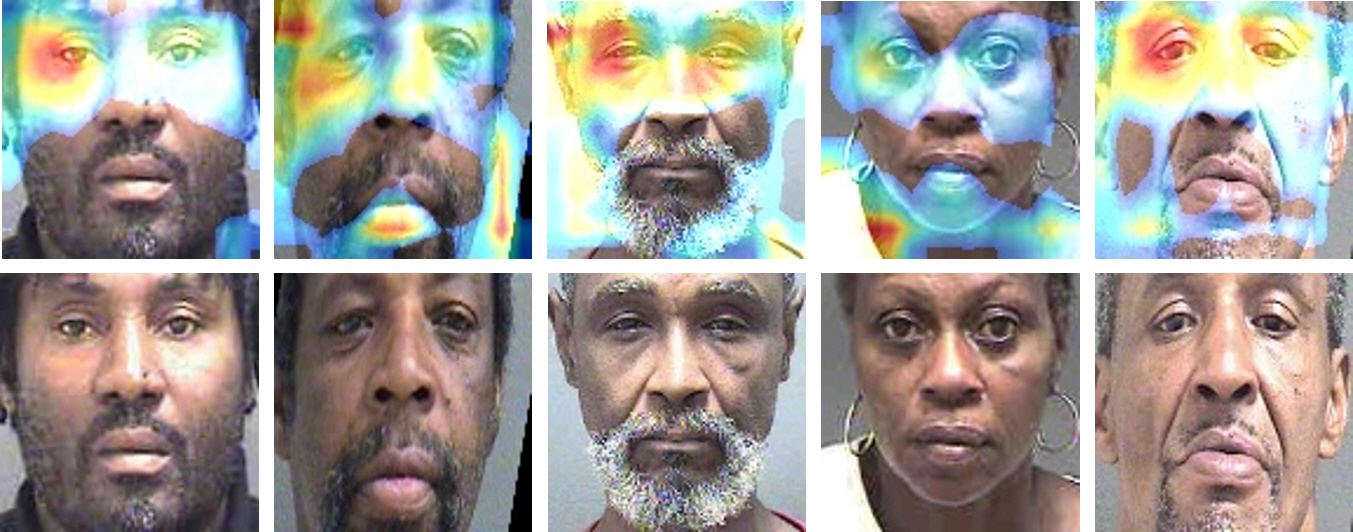}
    \caption{}
    \label{fig:figure_10c}
\end{subfigure}
\caption{CAMs of each expert gender classification from MORPH-\Romannum{2} (a) young, (b) adult, (c) elder.}
\label{fig:figure_10}
\end{figure}

Fig.~\ref{fig:figure_9}~and~\ref{fig:figure_10} show Class Activation Map (CAM) of each image to represent the discriminative facial regions to be used for the proposed gender classification expert~\cite{7780688}.~The images are divided into three groups based on their ground truth age (young, adult, elder), and each gender classification expert is used to generate CAMs of the corresponding images (e.g. CAM of young facial image is generated by young expert).

From Fig.~\ref{fig:figure_9}~and~\ref{fig:figure_10}, we can notice that each expert has different discriminative facial regions for the gender classification, and this corresponds to the idea that the facial feature representing gender is changed as person ages.~The highlighted regions for young and elder group are located over eye and mouth area.~This means that the discriminative feature for the gender classification of young and elder group is mainly acquired from eye and mouth area.~On the other hand, for the adult images, the highlighted regions are distributed over all facial area.~The CAMs of the young group images in MORPH-\Romannum{2} (Fig.~\ref{fig:figure_10a}) are similar to those of the adult group, because their ages are close to the adult group.~This result corresponds to the observation of Fig.~\ref{fig:figure_1} that the skin difference between male and female in the adult group is clear to be used for the gender classification but not enough for young and elder group.

\subsubsection{Performance of MGA}
The proposed MGA is designed to improve the gender classification accuracy by leveraging the result of age group classification with prediction of the gender classification experts.~The MGA is constructed as explained in section~\ref{Multi-expert Gender Classification on Age Group}, and test results are listed in Table~\ref{The age and gender classification accuracy of MGA on MORPH}~and~\ref{The age and gender classification accuracy of MGA on Adience}.

\begin{table}[!h]
\centering
\caption{The age estimation and gender classification accuracies of MGA on MORPH-\Romannum{2}.}
\label{The age and gender classification accuracy of MGA on MORPH}

\begin{adjustbox}{width=0.46\textwidth}
\begin{tabular}{p{3.5cm}llllll}
\hline
\multirow{2}{*}{Method} & \multicolumn{4}{l}{Gender accuracy (\%)} & \multirow{2}{*}{Age MAE} \\ \cline{2-5}
 & Young & Adult & Elder & Total &  \\ \hline
IN & 98.62 & 98.98 & 98.83 & 98.93 & 3.05 \\
MGA & \textbf{98.72} & \textbf{99.16} & \textbf{98.91} & \textbf{99.09} & \textbf{3.01} \\ \hline
\end{tabular}
\end{adjustbox}
\end{table}

\begin{table}[!h]
\centering
\caption{The age estimation and gender classification accuracies of MGA on Adience.}
\label{The age and gender classification accuracy of MGA on Adience}

\begin{adjustbox}{width=0.46\textwidth}
\begin{tabular}{p{2cm}llllll}

\hline
\multirow{2}{*}{Method} & \multicolumn{4}{l}{Gender accuracy (\%)} & \multicolumn{2}{l}{Age accuracy} \\ \cline{2-7} 
 & Young & Adult & Elder & Total & Exact (\%) & 1-off (\%) \\ \hline
IN & 87.80 & 94.01 & 89.93 & 90.94 & 73.86 & 87.57 \\
MGA & \textbf{88.80} & \textbf{94.91} & \textbf{91.30} & \textbf{91.93} & \textbf{75.03} & \textbf{87.80} \\ \hline
\end{tabular}
\end{adjustbox}
\end{table}

From the result, we can notice that the gender classification and age estimation accuracies are improved by adopting the MGA for all age groups.~Especially, this improvement is more significant for Adience dataset, because its age distribution is more diverse than MORPH-\Romannum{2}.~The classification accuracy is improved over than 1\% for young and elder group in Adience.~Therefore, the results support that the MGA is properly trained by the our training process and combines the result of age group classification and the prediction of gender classification expert to compensate the facial aging effects.

\subsubsection{MGA with ResNet}
As mentioned in Section~\ref{CAN}, the lightweight CNN is used for CAN to show superiority of the IN and MGA when they are applied for the gender and age estimation.~Thus, from the above tests (Table~\ref{CAN,DGN,IN on MORPH}$\sim$\ref{The age and gender classification accuracy of MGA on Adience}), we show that the proposed IN and MGA improve the performance of the gender classification and age estimation.

In this section, ResNet~\cite{DBLP:journals/corr/HeZRS15} is adopted for CAN to show the feasibility and superiority of the proposed MGA when deeper CNN architecture is applied.~Especially, 50-layer ResNet is used for the test, and the results are listed in table~\ref{Resnet50 on MORPH}~and~\ref{Resnet50 on Adience}.

\begin{table}[!h]
\centering
\caption{The age estimation and gender classification accuracies of MGA with ResNet-50 on MORPH-\Romannum{2}.}
\label{Resnet50 on MORPH}

\begin{adjustbox}{width=0.45\textwidth}
\begin{tabular}{p{4cm}lllllll}
\hline
Method & Gender (\%) & Age (MAE) \\ \hline
ResNet-50         & 98.85   &  2.95       \\
IN (ResNet-50)    &  98.93  & \textbf{2.93}        \\
MGA (ResNet-50)   &  \textbf{99.11}  & 2.95        \\ \hline

\end{tabular}
\end{adjustbox}
\end{table}

\begin{table}[!h]
\centering
\caption{The age estimation and gender classification accuracies of MGA with ResNet-50 on Adience.}
\label{Resnet50 on Adience}

\begin{adjustbox}{width=0.45\textwidth}
\begin{tabular}{p{3.5cm}lllllll}
\hline
\multirow{2}{*}{Method} & \multirow{2}{*}{Gender (\%)} & \multicolumn{2}{l}{Age} \\ \cline{3-4}
                        &       & Exact (\%)     & 1-off (\%)                  \\ \hline
ResNet-50                     &   92.85 &  76.82     &   87.95                     \\
IN (ResNet-50)                &    93.40 &  78.81    &   89.30                     \\
MGA (ResNet-50)               & \textbf{93.80}    &  \textbf{79.55}     & \textbf{89.74}    \\ \hline
\end{tabular}
\end{adjustbox}
\end{table}

In table~\ref{Resnet50 on MORPH}~and~\ref{Resnet50 on Adience}, ResNet-50 indicates the test result obtained by sole 50-layer ResNet.~From the result, we can notice that MGA (ResNet-50) achieves the best accuracy, which means that the proposed IN and MGA are feasible to the existing CNN architecture with better performance.~It is shown that the accuracy improvement is more significant for Adience because of its age distribution.

\subsubsection{Comparison with the-state-of-the-art results}
In this section, we compare the performance of the MGA with the state-of-the-art deep learning-based age estimation and gender classification methods.~The results are listed in table~\ref{Comparison with the-state-of-the-arts on MORPH}~and~\ref{Comparison with the-state-of-the-arts on Adience}.

\begin{table}[!h]
\centering
\caption{Comparison with the-state-of-the-arts on MORPH-\Romannum{2}.}
\label{Comparison with the-state-of-the-arts on MORPH}

\begin{adjustbox}{width=0.48\textwidth}
\begin{tabular}{p{3cm}llllll}

\hline
Method & Param \# & Gender (\%) & Age (MAE) \\ \hline
Heterogeneous~\cite{Han2018} & 61M & 98.0 & 3.0 \\
$Net_{Hybrid}^{VGG}$~\cite{Xing2017} & 138M & 98.7 & 2.96 \\
ResNet-50 & 23.6M & 98.85 & \textbf{2.95} \\
MGA & \textbf{2M} & 99.09 & 3.01 \\ 
MGA (ResNet-50) & 24M &  \textbf{99.11}  &  \textbf{2.95}\\\hline
 
\end{tabular}
\end{adjustbox}
\end{table}

\begin{table}[!h]
\centering
\caption{Comparison with the-state-of-the-arts on Adience.}
\label{Comparison with the-state-of-the-arts on Adience}

\begin{adjustbox}{width=0.48\textwidth}
\begin{tabular}{p{3cm}llllll}

\hline
\multirow{2}{*}{Method} & \multirow{2}{*}{Param \#} & \multirow{2}{*}{Gender (\%)} & \multicolumn{2}{l}{Age} \\ \cline{4-5} 
 &  &  & Exact (\%) & 1-off (\%) \\ \hline
 Attention~\cite{RODRIGUEZ2017563} & 138M & 93.0 & 61.8 & 95.10 \\
 RoR-34~\cite{Zhang2017}& 21.8M & 93.24 & 66.74 & \textbf{97.38} \\
 ResNet-50 & 23.6M & 92.85& 76.82& 87.95\\
 MGA & \textbf{2M} & 91.93 & 75.03 & 87.80 \\ 
 MGA (ResNet-50)    & 24M   &   \textbf{93.80}    &  \textbf{79.55}     & 89.74  \\\hline
 
\end{tabular}
\end{adjustbox}
\end{table}

For both MORPH-\Romannum{2} and Adience, the MGA achieves prominent results for gender classification and age estimation with only 2M parameters, which are much less than those of the previous methods (69 times smaller than \cite{RODRIGUEZ2017563, Xing2017}).~This result proves that the MGA is efficient to the gender classification due to compensating the facial aging effect.~In addition, the MGA also shows promising age estimation accuracy compared to the previous methods.

With ResNet-50, the MGA (ResNet-50) obtains the best gender classification and the age estimation accuracies, which outperform the previous methods.~This result shows that the proposed MGA is feasible to the previous deep learning-based gender classification and age estimation methods, and significant performance improvement is expected.

\section{Conclusion}
In this paper, we propose the multi-expert gender classifications on age groups to address the facial aging effect and improve the performance of the gender classification.~Each gender classification expert is constructed by integrating two networks: CAN and DGN.~The CAN is implemented to extract facial appearance feature, and the DGN is proposed for the facial geometric feature.~CAN and DGN are combined by the proposed model integration strategy, and the test results show that the IN achieves better performance, which means that the geometric feature is efficient for the deep learning-based gender classification when it is combined with CNN.

~The proposed MGA is composed of the age and gender parts.~The input image is categorized into one of three age groups (young, adult and elder), and the final gender prediction is made by the average fusion strategy of three gender classification experts trained to fit in each age group respectively.~The performance of the MGA is evaluated on MORPH-\Romannum{2} and Adience, and the results demonstrate that the MGA improves the gender classification and age estimation accuracy with small model size.~In addition, the experiments of MGA with ResNet-50 show that the MGA can be applied to the existing CNN architectures for better performance.~In our future work, we will investigate a novel approach to improve the age estimation accuracy using facial gender information. 

{\small
\bibliographystyle{ieee}
\bibliography{main}
}

\end{document}